\documentclass[10pt]{article}
\usepackage{tgpagella} 

\usepackage{tablefootnote}
\usepackage[dvips]{graphics}
\usepackage[utf8]{inputenc}
\usepackage{amsmath}
\usepackage{amsfonts}
\usepackage{amssymb}
\usepackage{graphicx}
\usepackage{fancyhdr}
\usepackage{multirow}
\usepackage{wrapfig, framed}
\usepackage{algorithm}
\usepackage{algorithmic}
\usepackage{epsfig,subfigure,endnotes,color,url,paralist, multirow,float}
\usepackage[left=2cm,right=2cm,top=2cm,bottom=2cm]{geometry}
\usepackage[inline,shortlabels]{enumitem}
\usepackage{mathtools}
\usepackage{mathrsfs}
\usepackage[table]{xcolor}
\definecolor{lightgray}{gray}{0.9}
\definecolor{LightCyan}{rgb}{0.88,1,1}

\usepackage{lipsum}
\usepackage{comment}
\usepackage{soul}
\usepackage{tikz}
\usepackage{xcolor}
\usepackage[format=plain,
            labelfont={bf,it},
            textfont=it]{caption}
\usepackage{xspace}
\usepackage{hyperref}
\usepackage{booktabs}

\usepackage{cite}
\usepackage{url}

\usepackage[compact]{titlesec}

\begin{document}

\newpage
\pagenumbering{arabic}
\pagestyle{plain}
\setcounter{page}{1}
\setcounter{section}{0}


\begin{center}{{\bf \LARGE
Making AI Less ``Thirsty'': Uncovering and Addressing the Secret Water Footprint of AI Models}}
\end{center}

\begin{table}[!h]
\centering
\begin{tabular}{m{0.2\textwidth} m{0.2\textwidth} m{0.2\textwidth} m{0.2\textwidth}}
\centering Pengfei Li\\ \emph{UC Riverside}
& \centering Jianyi Yang\\ \emph{UC Riverside} & \centering Mohammad A. Islam\\ \emph{UT Arlington} & \centering Shaolei Ren\tablefootnote{Corresponding author: Shaolei Ren (shaolei@ucr.edu), University of California, Riverside.}\\
 \emph{UC Riverside}
\end{tabular}
\end{table}

\begin{center}
    \textbf{Abstract}
\end{center}

The growing carbon footprint of artificial intelligence (AI)
 has been
undergoing public scrutiny. Nonetheless, the equally important
 water (withdrawal and consumption) footprint of AI has largely remained under the radar.
For example,
training the GPT-3 language model in Microsoft's state-of-the-art U.S. data centers
can directly evaporate {700,000 liters} of clean freshwater, but such information has been kept a secret. More critically, the global AI demand is projected to
account for {4.2 -- 6.6 billion cubic meters} of water withdrawal in 2027,
 which is more than the total
annual water withdrawal of {4 -- 6 Denmark} or
{half of the United Kingdom}. This is concerning, as
freshwater scarcity has become one of the most pressing challenges.
To respond
to the global water challenges,
AI
 can, and also must, take social responsibility and lead by example by addressing its own water footprint.
In this paper, we provide a principled methodology
to estimate the water footprint of AI, and also
discuss the unique spatial-temporal diversities of AI's
runtime water efficiency. Finally, we highlight the necessity
of holistically addressing water footprint along with carbon footprint
to enable truly sustainable AI.

\vspace{0.3cm}

\section{Introduction}

Artificial intelligence (AI) has enabled remarkable breakthroughs
in numerous areas of critical importance, including tackling global challenges such as climate change.
On the other hand, many AI models, especially large generative ones like GPT-4, are trained and deployed
on energy-hungry servers in warehouse-scale data centers, accelerating the data center energy consumption at an unprecedented rate \cite{DoE_DataCenter_EnergyReport_US_2024}. As a result,
AI's carbon footprint
has been undergoing scrutiny, driving the recent progress in AI carbon efficiency \cite{GreenAI_Washington_ACM_2020_10.1145/3381831,GreenAI_EnergyPolicy_NLP_UMass_ACL_2019_strubell-etal-2019-energy}. However, AI's water footprint
--- many millions of liters of freshwater consumed for
cooling the servers and for
electricity generation
 --- has largely remained under the radar and keeps escalating.
 If not properly addressed, AI's water footprint can potentially become a  major roadblock to sustainability
 and create social conflicts as freshwater resources suitable
 for human use are extremely limited and unevenly distributed.

As acknowledged in Google's sustainability report \cite{Google_SustainabilityReport_2024} and
the recent U.S. data center energy report \cite{DoE_DataCenter_EnergyReport_US_2024},
the expansion of
AI products and services is a key driver of
the rapid increase in data center water consumption.
Even excluding the water usage in leased third-party colocation facilities,
one technology company's self-owned
data centers alone directly withdrew 29 billion liters
and consumed (i.e., evaporated) more than 23 billion liters of freshwater for on-site cooling in 2023, nearly 80\% of which was potable water \cite{Google_SustainabilityReport_2024}.\footnote{The detailed difference between
water {withdrawal} and water {consumption}
is presented in Section~\ref{sec:water_usage_difference}.}
This amount of annual water consumption even rivals
that of a major household-name beverage company  \cite{PepsiCo_ESG_Water_24billion_2023}.
 Importantly, the company's data center water consumption
increased by
 $\sim$20\% from 2021 to 2022 and by $\sim$17\% from 2022 to 2023 \cite{Google_SustainabilityReport_2024}, and
 another technology company's data center water consumption
saw $\sim$34\% and $\sim$22\% increases over the same periods, respectively \cite{Microsoft_SustainabilityReport_2024}.
Furthermore, according to the recent U.S. data center energy report, the total annual on-site water consumption by U.S. data centers in 2028 could double or even quadruple the 2023 level, reaching approximately 150 -- 280 billion liters and further stressing
the water infrastructures \cite{DoE_DataCenter_EnergyReport_US_2024}.

AI represents the fastest expanding workloads in data centers \cite{Google_SustainabilityReport_2024,DoE_DataCenter_EnergyReport_US_2024}.
 For example, a recent study suggests that the global AI could
consume 85 -- 134 TWh of electricity in 2027 \cite{AI_Energy_Netherlands_2027_Joule_2023}, whereas
a more aggressive projection by the recent U.S. data center energy report predicts
that AI servers' electricity consumption in the U.S. alone will surpass 150 -- 300 TWh in 2028 \cite{DoE_DataCenter_EnergyReport_US_2024}.
Even considering the lower estimate,
the combined scope-1 and scope-2  water withdrawal of global AI is projected to reach \textbf{4.2 -- 6.6 billion cubic meters} in 2027, which is more than
the total annual water withdrawal of {4 -- 6 Denmark}
or {half of the United Kingdom}.\footnote{The scope
definition of water usage \cite{Water_Electricity_EWIF_Water_Intensity_WorkingPaper_WorldResourcesInstitute_2020_reig2020guidance} is in line with that of carbon
emissions and is discussed in Section~\ref{sec:water_usage_description}.
Our scope-2 water withdrawal (and consumption when applicable) is for location-based electricity generation throughout the paper.
Large data centers often adopt sustainability programs (e.g., renewable purchasing agreements) to offset their location-based electricity usage and thus may have lower market-based carbon and water footprints.}
Simultaneously, a total of 0.38 -- 0.60 billion cubic meters of water will be
evaporated and considered ``consumption'' due to the global
AI demand in 2027. Moreover, these global estimates will be exceeded
by the total water withdrawal and consumption attributed to AI in the U.S. alone in 2028 if the projection in \cite{DoE_DataCenter_EnergyReport_US_2024} comes to fruition.

Despite its profound environmental
and societal impact, the increasing water footprint of AI
has received disproportionately less attention from the AI community as well as the general public.
For example, while the scope-2 carbon emissions are routinely included as part of AI model cards, even scope-1 direct water usage (either withdrawal or consumption) is missing, let alone scope-2 water usage. This may impede innovations to enable water sustainability and build truly sustainable AI. Crucially, water and carbon footprints are complementary to, not substitutable of, each other for understanding the environmental impacts.
Indeed, optimizing for carbon efficiency does not necessarily
result in, and may even worsen, water efficiency, which varies with the
fuel mixes
for electricity generation and outside weather in a unique way \cite{Shaolei_Water_SpatioTemporal_GLB_TCC_2018_7420641,DoE_DataCenter_EnergyReport_US_2024}.

To ensure that the growth in AI does not exacerbate
the global water stresses or outweigh the environmental benefits it provides,
it is a critical time to uncover and address AI's hidden water footprint amid the increasingly severe freshwater scarcity crisis, worsened extended droughts, and quickly aging public water infrastructure. The urgency can also be reflected
in part by the recent commitment to ``{Water Positive by 2030}''
from industry leaders, including Google \cite{Google_SustainabilityReport_2024} and
Microsoft \cite{Microsoft_SustainabilityReport_2024},
and by the inclusion of water footprint as a key metric
into the world's first international standard on sustainable AI to be published by the ISO/IEC \cite{Water_StandardSustainableAI_ISOIEC}.

In this paper, we advocate for a holistic approach to sustainable AI that extends beyond the carbon footprint to also address the water footprint.
Specifically,
we present a principled methodology to estimate AI's total water footprint, including both operational
water and embodied water.  By taking the GPT-3 model with 175 billion parameters
as an example \cite{ML_GPT3_Energy_Others_NIPS_2020_NEURIPS2020_1457c0d6},
we show that
training GPT-3 in Microsoft's  U.S. data centers
can {consume} a total of {5.4 million liters} of water, including
\textbf{700,000 liters} of scope-1 on-site water consumption.
Additionally, GPT-3 needs to ``drink'' (i.e., consume) a \textbf{500ml bottle of water} for roughly 10 -- 50 medium-length responses, depending on
when and where it is deployed.

Next, we show that
WUE (Water Usage Effectiveness, a measure of water efficiency)
varies both spatially and temporally, suggesting
that judiciously deciding ``when'' and ``where'' to train a large AI model can significantly
cut the water footprint.
We also emphasize the need for increasing transparency
of AI' water footprint, including disclosing more information
about operational data and keeping users informed
of the runtime water efficiency.
 Finally, we highlight the necessity
of holistically addressing water footprint along with carbon footprint
to enable truly sustainable
AI --- \emph{the water footprint of AI can no longer stay under
the radar}.

\section{Background}

\subsection{Water Withdrawal vs. Water Consumption}\label{sec:water_usage_difference}

There are two related but different concepts --- water withdrawal and water consumption, both of which are important for understanding the impacts on water stress and availability  \cite{Water_Consumption_Withdrawal_WorldResourceInstitute,Water_EWIF_macknick2011review}. 

$\bullet$ \textbf{Water withdrawal:} It refers to
freshwater taken from the ground or surface water sources, either temporarily
or permanently, and then used for agricultural, industrial, or municipal uses (normally excluding water used for hydroelectricity generation) \cite{Water_Consumption_Withdrawal_WorldResourceInstitute}. As water is a finite shared resource,
water withdrawal
indicates the level of competition as well as dependence on water resources among different sectors.

$\bullet$ \textbf{Water consumption:}
It is defined
as ``water withdrawal minus water discharge'', and means
the amount of water ``evaporated, transpired, incorporated into products or crops, or otherwise removed from the immediate water environment'' \cite{Water_EWIF_macknick2011review}. Water consumption reflects the impact 
on downstream water availability and is crucial for assessing watershed-level scarcity \cite{Water_Consumption_Withdrawal_WorldResourceInstitute}.

These two types of water usage correspond to two different water footprints, i.e., water withdrawal footprint (WWF) \cite{Water_WaterWithdrawalFootprint_EnergySupply_Princeton_2024_https://doi.org/10.1111/jiec.12086,Water_Electricity_EWIF_Water_Intensity_WorkingPaper_WorldResourcesInstitute_2020_reig2020guidance} and 
water consumption footprint (WCF), respectively \cite{Water_DataCenterFootprint_EnvironmentalResearcHLetters_VT_2021_siddik2021environmental}. By default,
water footprint refers to the water consumption footprint unless otherwise specified. 

\subsection{How Does AI Use Water?}\label{sec:water_usage_description}

AI's water usage spans three scopes: on-site
water for data center cooling (scope 1), off-site water for
electricity generation (scope 2), and supply-chain water
for server manufacturing (scope 3).

\subsubsection{Scope-1 Water Usage}

Nearly all the server energy is converted into heat, which must then be removed
from the data center server room to avoid overheating. 
This process involves two sequential stages: server-level cooling followed by facility-level cooling.

In the server-level cooling stage, heat is transferred from the servers to the facility or a heat exchanger, typically using either air or liquid cooling methods (e.g., direct-to-chip cooling or immersion cooling), which do not evaporate or consume water. In general, new data centers dedicated to AI training often rely on liquid cooling due to the high server power densities.

In the facility-level cooling stage, heat
is rejected from the data center facility to the
outside environment. While there are various cooling
methods, 
 water-intensive cooling towers and water evaporation-assisted air cooling are two common approaches used in many data centers, including
 those operated by major technology companies \cite{Google_SustainabilityReport_2024,DoE_DataCenter_EnergyReport_US_2024}.

\begin{wrapfigure}[17]{r}{0.5\textwidth}
	\centering 
    \includegraphics[trim=0 4.3cm 3.8cm 0, clip, width=1\linewidth]{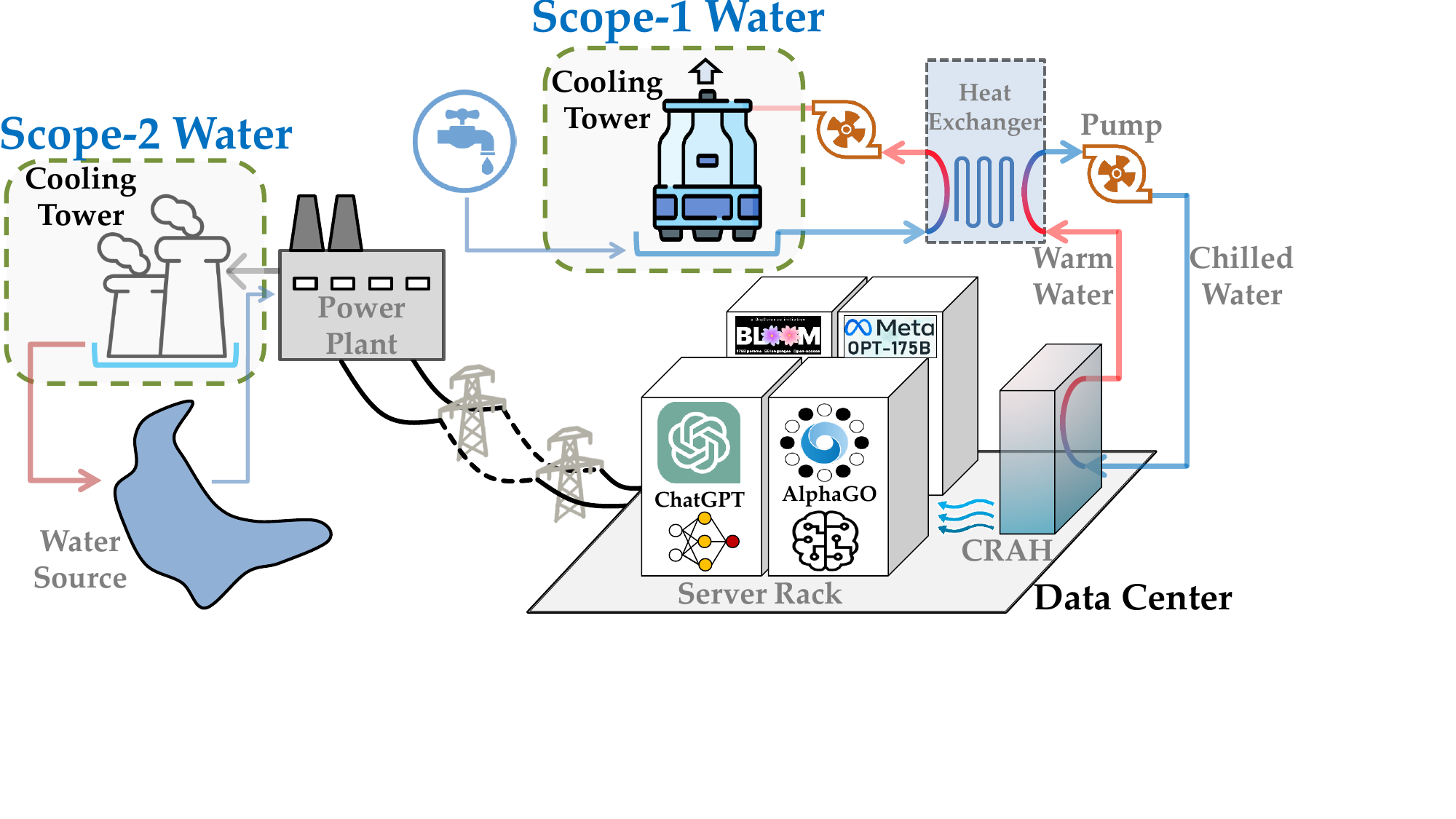}
\vspace{-0.5cm}    
\caption{An example of data center's operational water usage: on-site scope-1 water usage
for data center cooling (via cooling towers in the example), and off-site scope-2 water usage
for electricity generation. 
The icons for AI models are only for illustration purposes.} \label{fig:water_footprint}
	\vspace{-0.3cm}
\end{wrapfigure} 
\textbf{Cooling tower.} As illustrated in Figure~\ref{fig:water_footprint}, 
some water 
is evaporated (i.e., ``consumed'') in the cooling tower to dissipate heat into the environment, while the remaining water moves along an open loop to the heat exchanger to further absorb the server heat. Additionally, 
non-evaporated water can be recycled only a few times  (typically 3–10 cycles, depending on water quality) before discharge, requiring continuous clean freshwater replenishment to prevent mineral and salt buildup. 
Thus, to keep the cooling tower working, new water must be constantly added
to make up for the evaporated water and discharged water.
Importantly, clean freshwater (potable water in many cases \cite{Google_SustainabilityReport_2024}) is needed to avoid pipe clogs and/or bacterial growth.

For cooling towers, water withdrawal refers to the amount of added water, including both evaporated water and discharged water, while
water consumption exclusively indicates the amount of evaporated water. 
With good water quality, roughly 80\% of water withdrawal is evaporated and considered ``consumption'' \cite{Google_SustainabilityReport_2024}. 
  On average, depending on the
weather conditions and operational settings, data centers can
evaporate approximately 1 -- 9 liters per kWh of server energy: 1 L/kWh for Google's annualized global on-site water efficiency  \cite{Google_SustainabilityReport_2024} and 
9 L/kWh for a large commercial data center during the summer in Arizona
\cite{Water_DataCenterEnergy_Tradeoff_Arizona_Real_Measurement_WUE_Monthly_2022_KARIMI2022106194}.

\textbf{Air cooling with water evaporation assistance.} When the climate condition is appropriate, data centers may use ``free'' outside air to directly reject the heat to the outside environment. 
Nonetheless, water evaporation is still needed
when the outside air is too hot (e.g., higher than 85 degrees Fahrenheit);
additionally, water is also needed for humidity control when the outside air is too dry \cite{Facebook_Water_2023_meta}. The added water is considered ``withdrawal'', out of which about 
70\% is consumed based on Meta's report \cite{Facebook_SustainabilityReport_2024}.
Generally, outside air cooling is more water-efficient than cooling towers on average. However, hot weather raises the evaporative water demand and maximum water consumption, potentially stressing local water supplies during peak demand on hot days. 
 Additionally, the application of outside air cooling may have challenges in hot regions and/or
for many colocation facilities that are located in business districts.

Some data centers may opt for dry coolers, which consume no on-site water year-round \cite{Microsoft_Water_Zero_MoreEnergy_DataCenter_2024}. However, this approach typically increases cooling energy consumption compared to water-based cooling methods, potentially exacerbating the overall stress on water resources due to higher scope-2 water consumption.

\subsubsection{Scope-2 Water Usage}
In many countries, thermoelectric power is among
the top sectors in terms of water withdrawal
and water consumption \cite{Water_Electricity_EWIF_Water_Intensity_WorkingPaper_WorldResourcesInstitute_2020_reig2020guidance}.
Thus, similarly to scope-2 carbon emissions, data centers are accountable for off-site scope-2 water usage associated with electricity consumption, which forms part of the ``true water cost of data centers,'' as highlighted by the recent U.S. data
center energy report \cite{DoE_DataCenter_EnergyReport_US_2024}.

Different power plants use different amounts of water for each kWh generation, depending on the cooling techniques.
Typically, water withdrawal due to hydropower generation is excluded, but water consumption due to increased water evaporation rates from hydropower generation is included \cite{DoE_DataCenter_EnergyReport_US_2024}.
 For electricity generation,
the U.S. national average water withdrawal and consumption
are
 estimated at about 43.8 L/kWh \cite{Water_EnergyData_EIA_Website}
and 
3.1 L/kWh \cite{Water_Electricity_EWIF_Water_Intensity_WorkingPaper_WorldResourcesInstitute_2020_reig2020guidance}, respectively.
 Meta's self-reported 
scope-2 water consumption for its global data center fleet
was 3.7 L/kWh
(i.e., 55,475 megaliters divided by 14,975,435 MWh) in 2023
\cite{Facebook_SustainabilityReport_2024}.

\subsubsection{Scope-3 Water Usage}

AI chip and server manufacturing uses a huge amount
of water \cite{Water_LCA_SpatialTemporal_Semiconductor_2019_FROST2019100115,Water_Semiconductor_Singapore}. For example, ultrapure water is needed
for wafer fabrication and water is also needed for keeping
semiconductor plants cool. 
Importantly, the discharged water may contain toxic chemicals
and/or hazardous wastes. 
While water recycling at semiconductor plants can effectively
reduce water withdrawal, the recycling rate in many
cases remains low, e.g., the average
recycling rate for wafer plants and semiconductor plants
in Singapore are 45\% and 23\%, respectively 
\cite{Water_Semiconductor_Singapore}.
Although largely obscure, scope-3 water usage is likely significant \cite{Water_LCA_SpatialTemporal_Semiconductor_2019_FROST2019100115}. For instance, Apple reports that its supply chain accounts for 99\% of its total water footprint \cite{Apple_Sustainability_Report_2024}.

It is important to recognize that, unlike 
agriculture whose water footprint is mostly green
(i.e., water stored in soil and used by plants), the
majority of AI's water footprint is blue water
extracted from rivers, lakes, or groundwater, which is directly accessible for human use but often more limited in availability.
\section{Estimating AI's Water Footprint}

We present a general methodology
for estimating AI's water consumption footprint.
To obtain the water \emph{withdrawal} footprint, we simply
replace the WUE with
water withdrawal efficiency.

\subsection{Operational Water Footprint}\label{sec:methodology_operational}

We collectively refer to on-site scope-1 water and off-site scope-2 water
as the operational water.

$\bullet$ \textbf{On-site WUE.} We denote
the on-site scope-1 WUE at time $t$ by $\rho_{{s1},t}$, which
is defined as the ratio of the on-site water consumption to server energy
consumption and
varies over time depending on the outside temperature (see \cite{Shaolei_Water_SpatioTemporal_GLB_TCC_2018_7420641} for
an example of on-site WUE based on cooling towers).
Concretely, $\rho_{{s1},t}$ increases significantly for cooling towers
when the outside
wet bulb temperature  increases, and
increases for outside air cooling when the outside
dry bulb temperature is too hot or the humidity is too low.

$\bullet$ \textbf{Off-site WUE.}
We denote the off-site scope-2 WUE at time $t$ as $\rho_{s2,t}$, which
is defined as the ratio of off-site water consumption for each kWh of electricity
consumption and
measures the electricity water intensity factor (EWIF).
While there are different methods to estimate
 $\rho_{s2,t}$,
a common one is weighted averaging:
$\rho_{s2,t}
=\frac{\sum_k{b_{k,t}}\times EWIF_k}{\sum_k{b_{k,t}}}$
where $b_{k,t}$ denotes the amount of electricity generated from fuel type $k$ at time $t$ for the grid serving the
data center under consideration, and $EWIF_k$ is the EWIF for fuel type $k$ \cite{Shaolei_ICAC_2014_Water,Gao:2012:EG:2377677.2377719}.
Thus,
variations in energy fuel mixes of electricity generation result
in temporal variations of the off-site WUE. Moreover,
the off-site WUE also varies across regions due to
different energy fuel mixes
\cite{Water_Electricity_EWIF_Water_Intensity_WorkingPaper_WorldResourcesInstitute_2020_reig2020guidance,DoE_DataCenter_EnergyReport_US_2024}.

$\bullet$ \textbf{Operational water footprint.}
Consider a time-slotted model $t=1,2,\cdots,T$, where
the length of each time slot depends on how frequently
we want to assess the operational water footprint.
At time $t$, suppose that an AI model uses energy $e_t$ which can be measured using power meters and/or servers' built-in tools,
 and the data center hosting the AI model has a power
 usage effectiveness (PUE) of $\theta_t$ that accounts for the non-IT energy overhead.
Then, the total operational water footprint of the AI model can be written
as $WaterOperational = \sum_{t=1}^T e_t\cdot\left[ \rho_{s1,t}+
\theta_t\cdot \rho_{s2,t}\right]$.

\subsection{Embodied Water Footprint}\label{sec:methodology_embodied}

Similar to accounting for the embodied carbon footprint \cite{Carbon_LLM_Bloom_HuggingFace_2024_10.5555/3648699.3648952}, the total scope-3 water footprint is amortized over the lifespan of a server. Specifically, if $W$ represents the total water used to manufacture the AI servers and the servers are expected to operate for a period of $T_0$, then the embodied water footprint over a period of $T$ is calculated as $WaterEmbodied = \frac{T\cdot W}{T_0}$

By adding up the operational and embodied water footprints,
we can obtain the total water footprint as $WaterTotal = \sum_{t=1}^T e_t\cdot\left[ \rho_{s1,t}+
\theta_t\cdot \rho_{s2,t}\right]+\frac{T\cdot W}{T_0}$.
In practice, to obtain a rough estimate,
we can use the average values
for the annualized WUE and the estimated AI server energy
consumption.

\subsection{Case Study: Estimating GPT-3's Operational Water Consumption Footprint}\label{sec:example_gpt3}

\begin{table*}[!t]
\scriptsize
\centering
\caption{Estimate of GPT-3's operational {water consumption footprint}. ``*'' denotes data centers under construction as of July 2023, whose
  PUE and WUE are projected by Microsoft.}\label{Table:Estimated_Water_GPT3}
\begin{tabular}{c|c|c|c|c|c|c|c|c|c|c}
\toprule
\multirow{3}{*}{\textbf{Location}} & \multirow{3}{*}{\textbf{PUE}} & \multirow{3}{*}{\begin{tabular}[c]{@{}c@{}}\textbf{On-site}\\ \textbf{WUE}\\(L/kWh)\end{tabular}} & \multirow{3}{*}{\begin{tabular}[c]{@{}c@{}}\textbf{Off-site}\\ \textbf{EWIF}\\
(L/kWh)\end{tabular}} & \multicolumn{3}{c|}{\textbf{Water for Training} (million L)} & \multicolumn{3}{c|}{\textbf{Water for Each Request} (mL)}
& \multirow{2}{*}{\begin{tabular}[c]{@{}c@{}}\textbf{\# of Requests}\\~\textbf{for 500ml}~\\~\textbf{Water}\end{tabular}} \\
\cline{5-10}
&&&& \begin{tabular}[c]{@{}c@{}}On-site \\Water\end{tabular} & \begin{tabular}[c]{@{}c@{}}Off-site \\Water\end{tabular} & \begin{tabular}[c]{@{}c@{}}Total \\Water~\end{tabular} & \begin{tabular}[c]{@{}c@{}}On-site \\Water\end{tabular} & \begin{tabular}[c]{@{}c@{}}Off-site \\Water\end{tabular} & \begin{tabular}[c]{@{}c@{}}Total \\Water~\end{tabular}  \\
\hline
U.S. Average & 1.170 & 0.550 & 3.142& 0.708& 4.731& \textbf{5.439} & 2.200& 14.704 & \textbf{16.904} & \textbf{29.6}\\
\hline
Arizona~& 1.180 & 1.630 & 4.959& 2.098& 7.531& \textbf{9.629} & 6.520& 23.406 & \textbf{29.926}  &\textbf{16.7}\\
\hline
Georgia* & 1.120 & 0.060 & 2.309& 0.077& 3.328& \textbf{3.406} & 0.240& 10.345 & \textbf{10.585} & \textbf{47.2}\\
\hline
Illinois~& 1.350 & 0.740 & 2.233& 0.952& 3.880& \textbf{4.833} & 2.960& 12.060 & \textbf{15.020} & \textbf{33.3}\\
\hline
Iowa~ & 1.160 & 0.140 & 3.104& 0.180& 4.634& \textbf{4.814} & 0.560& 14.403 & \textbf{14.963} & \textbf{33.4}\\
\hline
Texas~& 1.280 & 0.250 & 1.287& 0.322& 2.120& \textbf{2.442} & 1.000& 6.590& \textbf{7.590} & \textbf{65.9}\\
\hline
Virginia~& 1.140 & 0.140 & 2.385& 0.180& 3.499& \textbf{3.679} & 0.560& 10.875 & \textbf{11.435} & \textbf{43.7}\\
\hline
Washington~ & 1.150 & 0.950 & 9.501& 1.223& 14.063 & \textbf{15.285} & 3.800& 43.706 & \textbf{47.506} & \textbf{10.5}\\
\hline
Wyoming~& 1.110 & 0.130 & 2.574& 0.167& 3.677& \textbf{3.845} & 0.520& 11.429 & \textbf{11.949} & \textbf{41.8}\\
\toprule
\hline
Australia*& 1.120 & 0.012 & 4.259& 0.015& 6.138& \textbf{6.154} & 0.048& 19.078 & \textbf{19.126} & \textbf{26.1}\\
\hline
Denmark* & 1.160 & 0.010 & 3.180& 0.013& 4.747& \textbf{4.760} & 0.040& 14.754 & \textbf{14.794} & \textbf{33.8}\\
\hline
Finland* & 1.120 & 0.010 & 4.542& 0.013& 6.548& \textbf{6.561} & 0.040& 20.350 & \textbf{20.390} & \textbf{24.5}\\
\hline
India* & 1.430 & 0.000 & 3.445& 0.000& 6.340& \textbf{6.340} & 0.000& 19.704 & \textbf{19.704} & \textbf{25.4}\\
\hline
Indonesia* & 1.320 & 1.900 & 2.271& 2.445& 3.858& \textbf{6.304} & 7.600& 11.992 & \textbf{19.592} & \textbf{25.5}\\
\hline
Ireland~& 1.190 & 0.020 & 1.476& 0.026& 2.261& \textbf{2.287} & 0.080& 7.027& \textbf{7.107} & \textbf{70.4}\\
\hline
Mexico*& 1.120 & 0.056 & 5.300& 0.072& 7.639& \textbf{7.711} & 0.224& 23.742 & \textbf{23.966} & \textbf{20.9}\\
\hline
Netherlands & 1.140 & 0.060 & 3.445& 0.077& 5.054& \textbf{5.131} & 0.240& 15.708 & \textbf{15.948} & \textbf{31.4}\\
\hline
Sweden & 1.160 & 0.090 & 6.019& 0.116& 8.986& \textbf{9.101} & 0.360& 27.927 & \textbf{28.287} & \textbf{17.7}\\
\bottomrule
\end{tabular}
\end{table*}

The core of ChatGPT, a popular online service, is
a large language model (LLM) based on subsequent versions
of GPT-3.
We present a case study to estimate the operational
water consumption for the full GPT-3 model with 175 billion parameters \cite{ML_GPT3_Energy_Others_NIPS_2020_NEURIPS2020_1457c0d6}.
We exclude embodied water footprint due to the lack of public
data for scope-3 water usage. 
We choose GPT-3 as Microsoft publishes its location-wise
WUE and PUE \cite{Microsoft_DataCenter_Community,Microsoft_DataCenter_PUE_WUE_Table_2023}.
 The results are summarized in Table~\ref{Table:Estimated_Water_GPT3}.

\subsubsection{Training}

GPT-3 was trained and deployed by OpenAI in Microsoft's data centers,
with an estimated training energy of 1287 MWh \cite{ML_Carbon_LargeModelTraining_Google_arXiv_2021_patterson2021carbon}.
 In line with the practice of estimating the carbon footprint, we use the most recent annualized average on-site PUE and WUE for each location, as reported by Microsoft  \cite{Microsoft_DataCenter_Community,Microsoft_DataCenter_PUE_WUE_Table_2023}.
 For power plant water efficiency, different references
may provide different estimates of
EWIF.
Thus, for consistency across regions, we use the EWIF provided by \cite{Water_Electricity_EWIF_Water_Intensity_WorkingPaper_WorldResourcesInstitute_2020_reig2020guidance} to estimate scope-2 water consumption, as it employs the same methodology for calculating EWIF. Moreover, a large number of Microsoft's data centers
are located in the U.S., where
the average EWIF provided by
\cite{Water_Electricity_EWIF_Water_Intensity_WorkingPaper_WorldResourcesInstitute_2020_reig2020guidance} is
3.14 L/kWh and significantly lower than
4.35 L/kWh reported by the recent
study
 \cite{DoE_DataCenter_EnergyReport_US_2024}.
 The specific location for training GPT-3 is not public.
Thus, we
consider Microsoft's different data center locations, while excluding Singapore and Taiwan as
EWIF data for these regions
is not available in \cite{Water_Electricity_EWIF_Water_Intensity_WorkingPaper_WorldResourcesInstitute_2020_reig2020guidance}.

\subsubsection{Inference}

As a representative usage scenario for an LLM, we consider a conversation task, which typically includes a CPU-intensive
prompt phase that processes the user's input (a.k.a., prompt)
and a memory-intensive token phase that produces outputs \cite{AI_DynamoLLM_ResourceManagement_LLM_Inference_energy_UIUC_Microsoft_ChaojieZhang_2024_stojkovic2024dynamollmdesigningllminference}.
More specifically, we consider a medium-sized request,
each with approximately $\leq$800 words of input and 150 -- 300 words of output \cite{AI_DynamoLLM_ResourceManagement_LLM_Inference_energy_UIUC_Microsoft_ChaojieZhang_2024_stojkovic2024dynamollmdesigningllminference}.
The official estimate indicates that GPT-3 consumes an order of 0.4 kWh of electricity to generate
100 pages of content, equivalent to roughly 0.004 kWh per page \cite{ML_GPT3_Energy_Others_NIPS_2020_NEURIPS2020_1457c0d6}.
While no details are provided, the estimate likely considers only the GPU energy used during token generation.

To account for both the prompt phase and the non-GPU energy consumption of servers, we assume a per-request server energy consumption of 0.004 kWh for our conversation task. The PUE, WUE, and EWIF are the
same as those used for estimating the training water consumption.
Our estimate of inference water consumption for GPT-3 is on the conservative side,
and the actual water consumption
 could be several times higher.
Specifically,
when considering service level objectives (SLOs) for LLM response times in enterprise-grade Nvidia DGX H100 systems for conversation tasks, the inference server energy consumption for a much smaller model (e.g., Llama-3-70B) is already approximately 0.010 kWh per medium-sized request when using a state-of-the-art LLM inference solution and accounting for non-GPU server overhead
\cite{AI_DynamoLLM_ResourceManagement_LLM_Inference_energy_UIUC_Microsoft_ChaojieZhang_2024_stojkovic2024dynamollmdesigningllminference}.
For the Falcon-180B model, which is comparable in size to GPT-3-175B, the server energy consumption reaches approximately 0.016 kWh per medium-sized request \cite{AI_DynamoLLM_ResourceManagement_LLM_Inference_energy_UIUC_Microsoft_ChaojieZhang_2024_stojkovic2024dynamollmdesigningllminference}.
Furthermore, we emphasize that Microsoft's data centers already have some of the lowest on-site WUE in the industry. If the same model is deployed in a third-party colocation data center, the scope-1 direct water consumption is expected to be several times higher. Additionally, our EWIF for the U.S. (3.14 L/kWh) is conservative and significantly lower than the 4.35 L/kWh recently reported by \cite{DoE_DataCenter_EnergyReport_US_2024}.

While no official information is available on the resource consumption,
some subsequent models like GPT-4 could consume substantially more energy and water than GPT-3 for processing the same
request  \cite{Shaolei_Water_Dataset_Africa_NeurIPS_Workshop_2024,ecologits-calculator}.
With continued efforts to reduce AI's
computational demand and improve the overall water efficiency, the water consumption per request may decrease in the
future. However,
 the total water consumption is likely to continue rising due to the growing demand for AI services and the increasing scale of AI applications \cite{DoE_DataCenter_EnergyReport_US_2024}.

\section{Our Recommendations}

We provide our recommendations to address
AI's water footprint from the scheduling and policy perspectives, making
future AI
more environmentally sustainable.

\begin{figure}[!t]
	\centering
      \subfigure[Carbon/water efficiency]{
\includegraphics[height=0.22\textwidth]{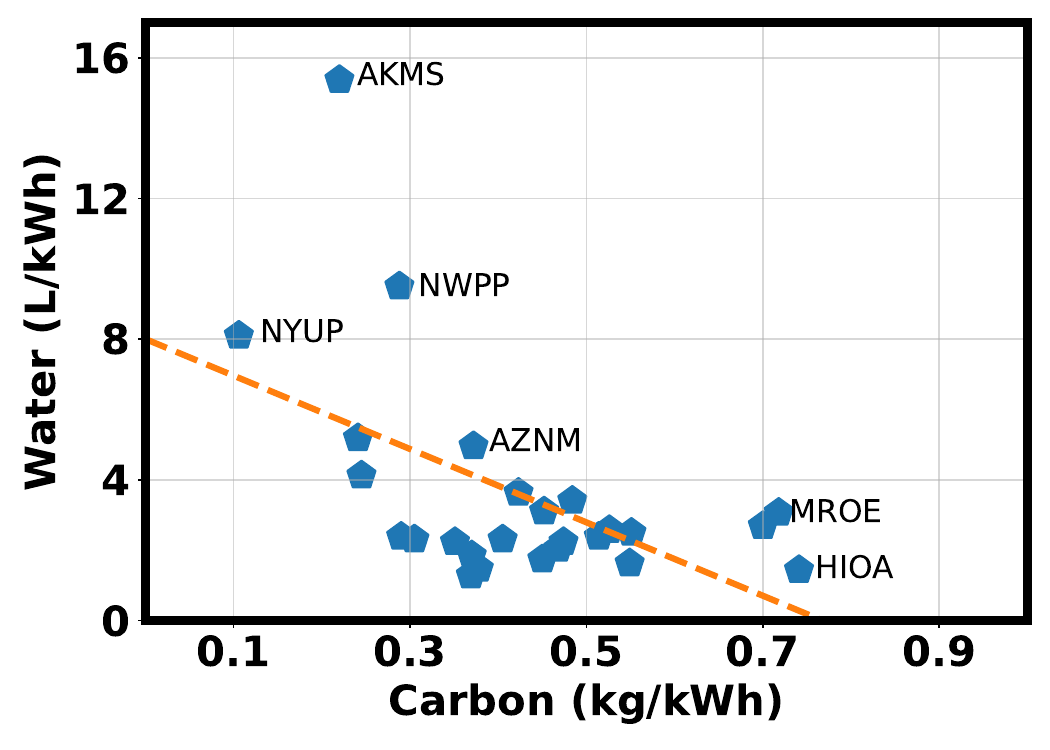}\label{fig:carbon_water_eGRID}}
     \subfigure[Hourly carbon/water efficiency]{\includegraphics[height=0.22\textwidth]{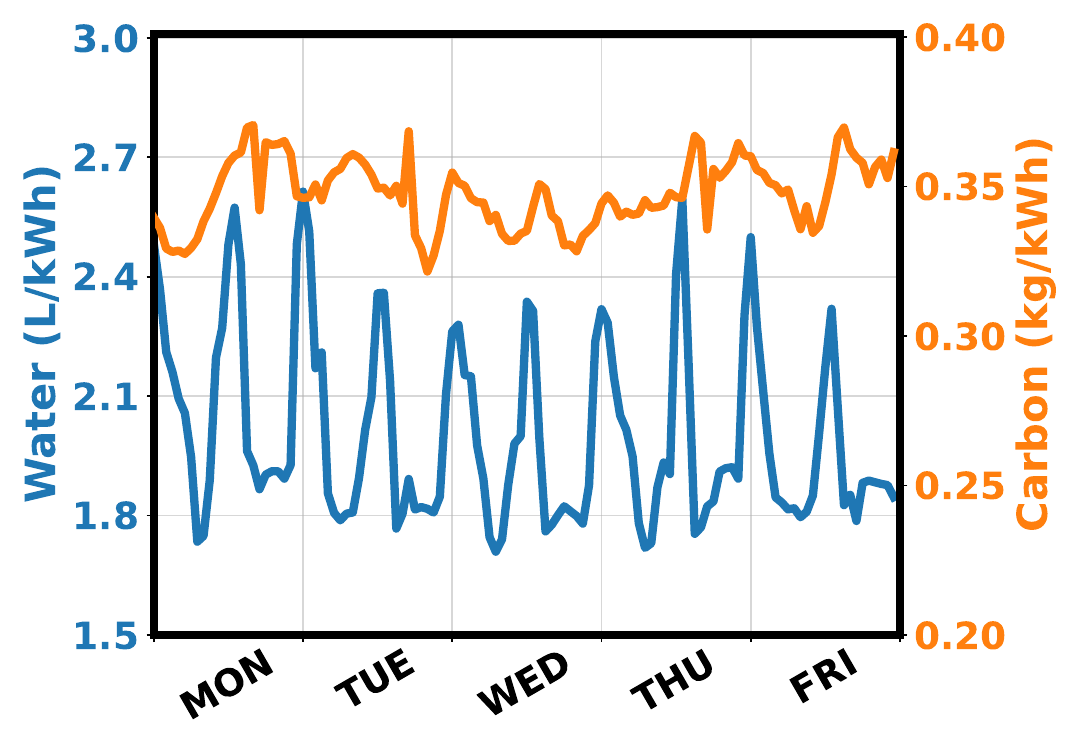}\label{fig:snapshot_carbon_water}
	}
	\subfigure[Hourly energy fuel mixes]{\includegraphics[height=0.22\textwidth]{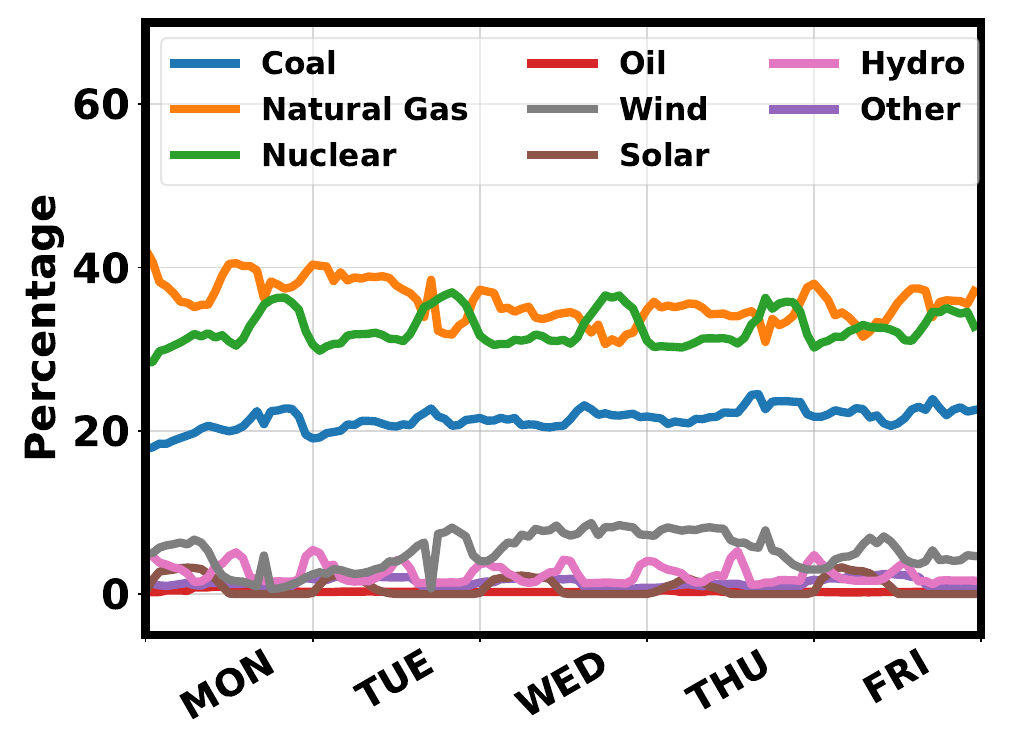}\label{fig:snapshot_fuel_mix}
	}%
 	\vspace{-0.3cm}	
	\caption{\textbf{(a)} The U.S. eGRID-level scope-2 water consumption intensity factor vs. carbon emission rate \cite{Carbon_eGRID_US_EPA_Website,Water_Electricity_EWIF_Water_Intensity_WorkingPaper_WorldResourcesInstitute_2020_reig2020guidance}. The dashed
 line represents a linear regression model, showing that the eGRID-level
 scope-2 carbon emission and water consumption efficiencies are not aligned.
 \textbf{(b)} A 5-day snapshot of scope-2 carbon emission rate and water consumption intensity
     in Virginia, starting from April 4, 2022. The values are calculated
     based on the fuel mixes, carbon emission rate
     and water consumption intensity for each fuel type \cite{Water_EnergyData_EIA_Website,Water_Electricity_EWIF_Water_Intensity_WorkingPaper_WorldResourcesInstitute_2020_reig2020guidance,Carbon_eGRID_US_EPA_Website}. The scope-2 carbon and water efficiencies only have a weak Pearson correlation coefficient of 0.06 in Virginia.
     \textbf{(c)} A 5-day snapshot of energy fuel mixes serving Virginia, starting from April 4, 2022 \cite{Water_EnergyData_EIA_Website}.}
     \label{fig:water_overall}
\end{figure}

\subsection{More Transparency and Comprehensive Reporting}

Despite its growing importance, AI's water footprint has received relatively less attention.
For example, while AI model cards routinely include carbon emissions
and serve as an important reporting framework for understanding
AI's environmental impacts, they currently omit information on AI's water consumption. The lack of transparency may obstruct efforts to drive innovations that enhance water sustainability and support truly sustainable AI. As an initial step to raise awareness among end users about the water resource impacts of their AI usage, we recommend tracking and reporting AI's water consumption in AI model cards and/or through cloud dashboards.

Moreover, a comprehensive understanding and reporting of AI's scope-2 water consumption associated with electricity generation remain limited. Although data centers have increasingly adopted climate-conscious cooling system designs to minimize on-site water consumption \cite{Google_SustainabilityReport_2024,Microsoft_Water_Zero_MoreEnergy_DataCenter_2024,Facebook_Water_2023_meta}, these efforts primarily focus on scope-1 water usage while largely overlooking scope-2 impacts.
Just as addressing scope-2 carbon emissions is important for mitigating climate change, it is equally crucial to address scope-2 water consumption to reduce AI's ``true water cost'',
as noted by
the recent U.S. data center energy report \cite{DoE_DataCenter_EnergyReport_US_2024}.
To better reflect the true impacts of data centers on water resources,
some technology companies such as Meta have begun to include scope-2 water consumption in their sustainability reports \cite{Facebook_SustainabilityReport_2024}.
We recommend the reporting of scope-2 water consumption as a standard practice. This approach makes the off-site water consumption visible to AI model developers as well as end users and can unlock new opportunities for demand-side flexibility,  thereby alleviating the overall strain on water resources.

Finally, despite the enormous scope-3 supply-chain water footprint \cite{Apple_Sustainability_Report_2024},
there is limited data available for embodied water usage by chip manufacturing.
We recommend further research on scope-3 water consumption to achieve a comprehensive understanding of AI's overall water footprint and to foster corporate water stewardship.

\subsection{``When'' and ``Where'' Matter}

Judiciously deciding ``when'' and ``where'' to train a large AI model can significantly
affect the water footprint.
The water efficiency exhibits a spatial-temporal diversity --- on-site water efficiency changes due to variations of outside weather conditions, and off-site water efficiency changes due to variations of the grid's energy fuel mixes to meet time-varying demands (Figure~\ref{fig:water_overall}).
Therefore,
we can dynamically schedule AI training and inference in a water-wise manner to
cut the water footprint. For example,
we may schedule AI training at midnight and/or
in a data center with better water efficiency.
Likewise, if informed of the real-time water efficiency, some water-conscious users may prefer to use
AI inference during water-efficient hours and/or in water-efficient
data centers, which can reduce
AI's water footprint by enabling demand-side flexibility.

\subsection{``Follow the Sun'' or ``Unfollow the Sun''}

To cut the carbon footprint, it is  preferable to ``follow
the sun'' when solar energy is more abundant. Nonetheless,
to cut the water footprint, it may be more appealing to
``unfollow the sun'' to avoid high-temperature hours
of a day when WUE is high.
This conflict
can also be shown in Figure~\ref{fig:carbon_water_eGRID} and Figure~\ref{fig:snapshot_carbon_water},
where we see
misalignment between the scope-2 water consumption intensity factor and
carbon emission
rate: minimizing one footprint might increase
the other footprint. This observation further corroborates
the previous finding that the environmental
impacts of carbon and water footprints are not substitutable \cite{DoE_DataCenter_EnergyReport_US_2024,Shaolei_Water_SpatioTemporal_GLB_TCC_2018_7420641}.
Therefore, to judiciously achieve a balance between
``follow
the sun'' for carbon efficiency
and ``unfollow
the sun'' for water efficiency, we need to reconcile
the potential water-carbon conflicts by using
 holistic approaches that are both carbon-efficient and water-wise.

\section{Conclusion}

In this paper, we uncover AI's water usage
as a critical concern for socially responsible and environmentally sustainable AI.
 We present a principled methodology to estimate AI's water footprint.
Then, using GPT-3 as an example, we show that a large AI model can consume
  millions of liters of water for training. We also
discuss that the scope-1 and scope-2 water efficiencies
vary spatially and temporally ---
 judiciously deciding ``when'' and ``where'' to run a large AI model can significantly
cut the water footprint.
In addition, we recommend increased transparency and comprehensive reporting
of AI's water footprint, and highlight the necessity
of holistically addressing the water footprint along with the carbon footprint
to build truly sustainable AI.

\emph{AI's water footprint can no longer stay under the radar and must be addressed as a priority as part of the collective
efforts to combat global water challenges.}

{
       \bibliographystyle{unsrt}

}

\newpage
\appendix 

\section*{Appendix: Operational Water for Global AI in 2027}

A recent study suggests that the global AI could
consume 85 -- 134 TWh of electricity in 2027 based on the GPU shipment \cite{AI_Energy_Netherlands_2027_Joule_2023}, whereas
a more aggressive estimate by the U.S. data center energy report projects
that AI servers' electricity consumption in the U.S. alone will surpass 150 -- 300 TWh in 2028 \cite{DoE_DataCenter_EnergyReport_US_2024}.
Based on the former and more conservative projection, we estimate the potential water usage for global AI in 2027, while noting
that our global estimates will be exceeded
by the water usage attributed to AI in the U.S. alone in 2028 if the  projection in \cite{DoE_DataCenter_EnergyReport_US_2024} comes to fruition.

\textbf{Scope-1 water usage.} The scope-1 water efficiency depends on a variety of factors, including
the cooling system designs, climate conditions, and operational settings. 
To set the global scope-1 water efficiency, we utilize the annualized water efficiencies reported by two leading data center operators, Google and Equinix, in their latest sustainability reports \cite{Google_SustainabilityReport_2024,Equinix_EnvironmentalSustainabilityReport_2024}. Specifically, for on-site scope-1 water withdrawal, we assume
1.2 L/kWh, which results in a total scope-1 water withdrawal of 0.11 -- 0.16 billion cubic meters. Similarly, assuming 1.0 L/kWh
for global scope-1 water consumption efficiency,
we obtain a total on-site scope-1 water consumption
of 0.09 -- 0.14 billion cubic meters.
Note that Google and Equinix both operate data centers globally, but represent two distinct categories of data centers: hyperscale data centers (Google) and multi-tenant colocation data centers (Equinix). According to the recent U.S. data center energy report \cite{DoE_DataCenter_EnergyReport_US_2024}, these two types of data centers collectively account for the vast majority of data center energy consumption in the U.S., with colocation data centers consuming slightly more energy than hyperscalers.

\textbf{Scope-2 water usage.}
As noted by the recent U.S. data center energy report \cite{DoE_DataCenter_EnergyReport_US_2024}, scope-2 water usage
is part of the true water cost of data centers. 
The U.S. average electricity water withdrawal and consumption intensity
factors are both lower than the global averages \cite{Water_Electricity_EWIF_Water_Intensity_WorkingPaper_WorldResourcesInstitute_2020_reig2020guidance}.
Thus, 
in our estimate,
 we use the U.S. average electricity water withdrawal intensity factor 43.83 L/kWh \cite{Water_WaterWithdrawal_US_Electricity_2022},
and electricity water consumption intensity factor 3.14 L/kWh \cite{Water_Electricity_EWIF_Water_Intensity_WorkingPaper_WorldResourcesInstitute_2020_reig2020guidance}, respectively.
Note that, since \cite{Water_Electricity_EWIF_Water_Intensity_WorkingPaper_WorldResourcesInstitute_2020_reig2020guidance} includes hydropower in the
calculation, it has a higher electricity water withdrawal factor
than the U.S. Energy Information Administration's calculation
(i.e., 386.07 L/kWh vs. 43.83 L/kWh for the U.S.).
Moreover, our value of 3.14 L/kWh for the U.S.
average water consumption factor is lower than 4.35 L/kWh reported
by  \cite{DoE_DataCenter_EnergyReport_US_2024},
as well as lower than Meta's global
electricity water consumption intensity factor of 3.70 L/kWh in 2024
(i.e., 55,475 megaliters divided by 14,975,435 MWh) 
\cite{Facebook_SustainabilityReport_2024}. 
Therefore, 
 our choices of 43.83 L/kWh and 3.14 L/kWh for electricity water withdrawal and consumption
intensity factors are both on the conservative side, which can partly absorb potential over-estimates of global AI's energy demand in 2027 provided by \cite{AI_Energy_Netherlands_2027_Joule_2023}.

To account for the data center non-IT energy overheads, we conservatively assume a power
usage effectiveness (PUE) of 1.1, which is a fairly low value even for 
state-of-the-art data center facilities \cite{Google_SustainabilityReport_2024}.
Thus, AI's total electricity consumption becomes 93.5 -- 147.4 TWh.
Thus, after multiplying 43.83 L/kWh and 3.14 L/kWh by 93.5 -- 147.4 TWh, we obtain the total scope-2 water withdrawal of 4.10 -- 6.46 billion cubic meters and water consumption of 0.29 -- 0.46 billion cubic meters, respectively.

\textbf{Total water usage.}
By adding up scope-1 and scope-2 water usage together,
the total water withdrawal and water consumption of global AI 
may reach 4.2 -- 6.6 billion cubic meters
and 0.38 -- 0.60 billion cubic meters, respectively.
According to the U.S. Central Intelligence Agency \cite{Water_WaterWithdrawal_Country_2020_CIA}, the estimated U.S. annual water withdrawals in Denmark and the United Kingdom in 2020 (the latest year available as of January, 2025) were 0.98 billion cubic meters
and 8.42 billion cubic meters, respectively.
Thus, assuming that the 2027 water withdrawals in these two countries
remain similar to their 2020 levels,
 the total water withdrawal attributed to global AI in 2027 is projected to surpass the equivalent of the total annual water withdrawal of 4 -- 6 Denmark or approximately half of the United Kingdom. The U.S. Central Intelligence Agency \cite{Water_WaterWithdrawal_Country_2020_CIA} does not provide the country-wide annual water consumption information, and hence we do not contextualize the total water consumption of global AI in 2027.

The estimates of global AI's water usage in 2027 are naturally subject
to uncertainties, e.g.,
 the future water efficiency may differ from the current value we use. 
Nonetheless, we emphasize that our estimates are on the conservative side. For example, according to the U.S. data center energy report, the scope-1 water consumption attributed to AI in the U.S. alone could exceed 0.2 billion cubic meters in 2028 \cite{DoE_DataCenter_EnergyReport_US_2024}. Moreover, based on the reported scope-2 water consumption efficiency, the combined scope-1 and scope-2 water consumption 
attributed to AI in the U.S. alone is projected to reach up to about 2 billion cubic meters in 2028 \cite{DoE_DataCenter_EnergyReport_US_2024},
which is significantly higher than our estimate of
global AI's total water consumption in 2027.

\end{document}